\documentclass[11pt,twoside]{article}
\usepackage{inf_neutral}
\usepackage{epsfig}

\usepackage[dvipsnames]{xcolor}
\usepackage{hyperref}
\usepackage{graphicx}

\colorlet{mylinkcolor}{Black}
\colorlet{mycitecolor}{Black}
\colorlet{myurlcolor}{Blue}
\usepackage{graphicx}
\usepackage{cite}

\hypersetup{
  linkcolor  = mylinkcolor,
  citecolor  = black,
  urlcolor   = myurlcolor,
  colorlinks = true,
}

\begin{document}
  \title{Robust Computer Algebra, Theorem Proving, and Oracle AI}
   \author{Gopal P. Sarma \\
      School of Medicine, Emory University, Atlanta, GA USA\\
      gopal.sarma@emory.edu \\
      \AndAuthor
      Nick J. Hay \\
      Vicarious FPC, San Francisco, CA USA\\
      nnickhay@gmail.com}
  \titleodd{Robust Computer Algebra, Theorem Proving, and Oracle AI}
  \authoreven{Sarma and Hay}
  \keywords{Oracle AI, AI safety, CAS, theorem proving, math oracles}
  \received{June 24, 2013}
\abstract{
In the context of superintelligent AI systems, the term ``oracle'' has two meanings.  One  
refers to modular systems queried for domain-specific tasks.  Another usage, referring to a class of 
systems which may be useful for addressing the value alignment and AI control problems, is a 
superintelligent AI system that only answers questions.  The aim of this manuscript is to survey 
contemporary research problems related to oracles which align with long-term research goals of AI 
safety.  We examine existing question answering systems and argue that their high degree of 
architectural heterogeneity makes them poor candidates for rigorous analysis as oracles.  On the other 
hand, we identify computer algebra systems (CASs) as being primitive examples of domain-specific oracles 
for mathematics and 
argue that efforts to integrate computer algebra systems with theorem provers, systems which have 
largely been developed independent of one another, provide a concrete set of problems related to the 
notion of provable safety that has emerged in the AI safety community.  We review approaches to 
interfacing CASs with theorem provers, describe well-defined architectural deficiencies that have been 
identified with CASs, and suggest possible lines of research and practical software projects for 
scientists interested in AI safety.  
}

\abstractSi{}

\maketitle

\section{Introduction} \label{introduction}
Recently, significant public attention has been drawn to the consequences of achieving human-level 
artificial intelligence.  While there have been small communities analyzing the long-term impact of AI 
and related technologies for decades, these forecasts were made before the many recent 
breakthroughs that have dramatically accelerated the pace of research in areas as diverse as robotics, 
computer vision, and autonomous vehicles, to name just a few \cite{bostrom2014superintelligence, 
shanahan2015technological, chalmers2010singularity}. \\

Most researchers and industrialists view advances in artificial intelligence as having the potential to be 
overwhelmingly beneficial to humanity.  Medicine, transportation, and fundamental scientific research 
are just some of the areas that are actively being transformed by advances in artificial intelligence.  On 
the other hand, issues of privacy and surveillance, access and inequality, or economics and policy are 
also of utmost importance and are distinct from the specific technical challenges posed by most 
cutting-edge research problems \cite{tegmark2015open, russell2015research}.  \\

In the context of AI forecasting, one set of issues stands apart, namely, the consequences of artificial 
intelligence whose capacities vastly exceed that of human beings.  Some researchers have argued that 
such a ``superintelligence''  poses distinct problems from the more modest AI systems described above.  
In particular, the emerging discipline of AI safety has focused on issues related to the potential 
consequences of mis-specifying goal structures for AI systems which have significant capacity to exert 
influence on the world.  From this vantage point, the fundamental concern is that deviations from 
``human-compatible values'' in a superintelligent agent could have significantly detrimental 
consequences \cite{bostrom2014superintelligence}.  \\

One strategy that has been advocated for addressing safety concerns related to superintelligence is 
Oracle AI, that is, an AI system that only answers questions.  In other words, an Oracle AI does not 
directly influence the world in any capacity except via the user of the system.  Because an Oracle AI 
cannot directly take physical action except by answering questions posed by the system's operator, 
some have argued that it may provide a way to bypass the immediate need for solving the ``value 
alignment problem'' and would itself be a powerful resource in enabling the safe design of autonomous, 
deliberative superintelligent agents \cite{armstrong2012thinking, bostrom2014superintelligence, 
fallenstein2015reflective, armstrong2017, armstrong2017oracle}.  \\

A weaker notion of the term oracle, what we call a \emph{domain-specific oracle}, refers to a modular 
component of a larger AI system that is queried for domain-specific tasks.  In this article, we view 
computer algebra systems as primitive domain-specific oracles for mathematical 
computation which are likely to become quite powerful on the time horizons on which many expect 
superintelligent AI systems to be developed \cite{muller2016future, 2017arXiv170508807G}.  
Under the assumption that math oracles prove to be 
useful in the long-term development of AI systems, addressing well-defined architectural 
problems with CASs and their integration with interactive theorem provers provides a concrete 
set of research problems that align with long-term issues in AI safety.  In addition, such systems may also
be useful in proving the functional correctness of other aspects of an AI architecture.  In Section 
\ref{metascience}, we briefly discuss the unique challenges in allocating resources for AI safety 
research.  In Section \ref{oracle-overview}, we briefly summarize the motivation for developing
oracles in the context of AI safety and give an overview of safety risks and control strategies 
which have been identified for superintelligent oracle AIs.  
In Section \ref{oracle} we analyze contemporary question answering systems and argue that 
in contrast to computer algebra systems, current consumer-oriented, NLP-based systems are poor 
candidates for rigorous analysis as oracles.  In Section \ref{itp-cas}, we review the differences between 
theorem provers and computer algebra systems, efforts at integrating the two, and known architectural 
problems with CASs.  We close with a list of additional research projects related to mathematical computation 
which may be of interest to scientists conducting research in AI safety.  

\section{Metascience of AI Safety Research}\label{metascience}
From a resource allocation standpoint, AI safety poses a unique set of challenges.  Few areas of 
academic research operate on such long and potentially uncertain time horizons.  This is not to say that 
academia does not engage in long-term research.  Research in quantum gravity, for example, is 
approaching nearly a century's worth of effort in theoretical physics \cite{Rovelli:2008}.  However, the 
key difference between open-ended, fundamental research in the sciences or humanities and AI safety 
is the possibility of negative consequences, indeed significant ones, of key technological 
breakthroughs taking place without corresponding advances in frameworks for safety 
\cite{bostrom2014superintelligence, russell2016should} .  \\

These issues have been controversial, largely due to disagreement over the time-horizons for achieving 
human-level AI and the subsequent consequences \cite{muller2016future, 2017arXiv170508807G}.  
Specifically, the notion of an ``intelligence explosion,'' whereby the intelligence of software systems 
dramatically increases due their capacity to model and re-write their own source code, has yet to receive 
adequate scientific scrutiny and analysis \cite{linstone2014singularity}. \\

We affirm the importance of AI safety research and also agree with those who have cautioned against 
proceeding down speculative lines of thinking that lack precision.  
Our perspective in this article is that it is 
possible to fruitfully discuss long-term issues related to AI safety while maintaining a connection to 
practical research problems.  To some extent, our goal is similar in spirit to the widely discussed 
manuscript ``Concrete Problems in AI Safety'' \cite{amodei2016concrete}.  However, we aim to be a bit 
more bold.  While the authors of ``Concrete Problems'' state at the outset that their analysis will set 
aside questions related to superintelligence, our goal is to explicitly tackle superintelligence related 
safety concerns.  We believe that there are areas of contemporary research that overlap 
with novel ideas and concepts that have arisen among researchers who have purely focused on 
analyzing the consequences of AI systems whose capacities vastly exceed those of human beings. \\

To be clear, we do not claim that the strategy of searching for pre-existing research objectives that align
with the aims of superintelligence theory is sufficient to cover the full spectrum of issues identified by 
AI safety researchers.  There is no doubt that the prospect of superintelligence raises entirely new 
issues that have no context in contemporary research.  However, considering how young the field is, 
we believe that the perspective adopted in this article is a down-to-earth and moderate stance to take 
while the field is in a critical growth phase and a new culture is being created. \\

This article focuses on one area of the AI safety landscape, Oracle AI.   We identify a set of concrete software 
projects that relate to more abstract, conceptual ideas from AI safety, to bridge the gap between 
practical contemporary challenges and longer term concerns which are of an uncertain time horizon.  
In addition to providing concrete problems for researchers and engineers to tackle, we hope
this discussion will be a useful introduction to the concept of Oracle AI for newcomers to the subject.  
We state at the outset that within the context of Oracle AI, our analysis is limited in scope to systems 
which perform mathematical computation, and not to oracles in general.  Nonetheless, considering how 
little effort has been directed at the 
superintelligence control problem, we are confident that there is low-hanging fruit in addressing these 
more general issues which are awaiting discovery.  

\section{Brief Overview of Oracle AI}\label{oracle-overview}
As described above, an Oracle AI is a system which only answers questions.  Although the term has traditionally
been used to denote fully general AI systems, we can also imagine oracles whose capabilities are more limited.  
The latter systems are what we refer to as \emph{domain-specific oracles}.  AI safety researchers have 
argued that superintelligent oracles have the potential to help address safety concerns of autonomous, 
deliberative superintelligent AI systems, as an Oracle AI's only ability to influence the world is through 
the user of the system.  Consequently, an Oracle AI itself may be in a position to assist researchers in
solving difficult technical and conceptual problems related to the design of independent, deliberative agents. 
Yet superintelligent oracle AIs pose their own distinct set of safety concerns.  
The problems identified in \cite{armstrong2012thinking}, for example, presume a superintelligent AI 
system, which while limited in its ability to interact with the world, may nonetheless have the potential to 
negatively impact the world by manipulating the user of the system.  As Armstrong, Bostrom, and Sandberg write:

\begin{quote}
{\small
We are all vulnerable to the OAI's social engineering. An advanced OAI can literally offer us the world. 
Genuine altruism is no defence against this; it can offer a world free of poverty, hunger, war, disease or 
death, as well as all personal trinkets we could want. It could guess the weaknesses of each individual, 
and find the right arguments to convince us that granting it power or liberty is the moral and profitable 
thing to do. It will be able to play on rivalries, institutional weaknesses and ethical dilemmas with more 
virtuosity than any human could. \cite{armstrong2012thinking}
}
\end{quote}

Given the potential dangers of human manipulation by an OAI, researchers have searched for control strategies
to mitigate such risks.  A few examples of control methods that have been proposed are:
\begin{itemize}
\item \textbf{Physical security}- physically containing the  OAI system
\item \textbf{Limiting or eliminating external manipulators}- like the previous security measure, 
precluding the OAI from physically manipulating the world in any direct manner
\item \textbf{Epistemic capability control}- the OAIs output could be restricted even further beyond 
question answering, for example, restricting it to answer ``yes / no '' questions
\item \textbf{Virtual worlds}- the OAI could be run in a simulated world so that even if the system's 
capabilities were expanded, its behavior could be observed and modeled.  
\item \textbf{Resetting the OAI after each question}- this would prevent the OAI from engaging in long-term 
social engineering by manipulating the answers to the users' questions
\end{itemize}

Although the capacities of domain-specific oracles are more limited, they can nonetheless pose safety risks. 
Architectural deficiencies in such oracles might be exploited by a larger AI system to 
manipulate the human user.  It could give answers which are difficult to verify and which allow the OAI to 
execute complex and intricate plans unbeknownst to the user.  Therefore, while flaws in domain-specific oracles are not
inherently risky if used solely in their domain of applicability, they may very well be dangerous as part of 
a larger system with more general capabilities.  Though not a ``control strategy'' in the narrowest sense, 
creating ``robust'' domain-specific oracles is an
important objective in designing safe OAIs.  Furthermore,
ensuring the robustness of domain-specific subsystems might mitigate the need for stronger control strategies,
as the OAI would have fewer weaknesses to exploit.  \\

It should go without saying that the arguments presented above are highly schematic and do not dependent
on specific technologies.  To our knowledge, there is very limited work on translating analyses of 
superintelligent oracle AIs into the concrete language of modern artificial intelligence 
\cite{armstrong2016safely, armstrong2017, armstrong2017oracle}.  Our goal in this manuscript is in this spirit, that is, to
anchor schematic, philosophical arguments in practical, contemporary research.  To do so, we will narrow our focus
to the mathematical domain.  In the remainder of the article, we will use the 
term oracle in the more limited sense of a domain-specific subsystem, and in particular, oracles for performing 
mathematical computations.  We hope that the analysis presented here will be of intrinsic value in 
developing robust math oracles, as well as provide some intuition and context for identifying 
concrete problems relevant to developing safe, superintelligent oracle AI systems.  

\section{Are there contemporary systems which qualify as oracles?}\label{oracle}
The obvious class of contemporary systems which would seem to qualify as oracles are question 
answering systems (QASs).  As we stated above, a basic criterion characterizing oracles is that their 
fundamental mode of interaction is answering questions posed by a user, or for domain-specific queries 
as part of a larger AI system. \\

Contemporary QASs are largely aimed at using natural language processing techniques to answer 
questions pertaining to useful facts about the world such as places, movies, historical figures, and so 
on.  An important point to make about QASs is the highly variable nature of the underlying technology.  
For instance, IBM's original Watson system which competed in Jeopardy, was developed prior to the 
recent advances in deep learning which have fundamentally transformed areas ranging from computer 
vision, to speech recognition, to natural language processing \cite{ferrucci2010building}.  In this 
particular task, the system was nonetheless able to perform at a level beyond that of the most 
accomplished human participants.  The introduction of ``info panes'' into popular search 
engines, on the other hand, have been based on more recent machine learning technology, and indeed, 
these advances are also what power the latest iterations of the Watson system 
\cite{watson_upgrade}.  On the other end of the spectrum is Wolfram $\vert$ Alpha, which is also a question 
answering system, but which is architecturally centered around a large, curated repository of structured 
data, rather than datasets of unstructured natural language \cite{wolfram_QAS}.  \\

While these systems are currently useful for humans in navigating the world, planning social outings, 
and arriving at quick and useful answers to ordinary questions, it is not clear that they will remain useful 
in quite the same capacity many years from now, or as standalone components of superintelligent AI 
systems.  Although the underlying techniques of deep learning or NLP are of fundamental interest in 
their own right, the fact that these systems are QASs at all seems to be more of an artifact of their utility for 
consumers.  \\

Another important observation about contemporary QASs is that much of their underlying NLP-based 
architecture can be replaced by taking advantage of structured data, as the example of Wolfram | Alpha 
demonstrates.  For the other NLP or machine learning based 
systems, the underlying technology can be used as part of larger, semi-automated pipelines to turn 
unstructured data from textual sources into structured data.  Once again, this fact simply underscores 
that contemporary QASs are not particularly appealing model systems to analyze from the Oracle AI 
safety perspective.\footnote{We emphasize that our argument that
contemporary QASs are not good candidates for analysis as Oracle AIs is not an argument 
against the traditional formulation of Oracle AI as a tool for AI safety.  We fully expect significant 
breakthroughs to be made in advancing the theory and practice of oracle-based
techniques for AI safety and we hope that this manuscript will provide some motivation 
to pursue such research.  Rather, our point is that when viewing
contemporary systems from the lens of superintelligence, there seems little reason to believe that 
current NLP-based QASs will remain sufficiently architecturally stable to be used as standalone components 
in AI systems many years from now.  On the other hand, there are certainly important \emph{present-day} problems 
to examine when evaluating the 
broader impact of QASs, such as bias in NLP systems, overgeneralization, and privacy, to name just a 
few.  Some of these issues overlap with the set of problems identified in \cite{amodei2016concrete} as 
examples of concrete problems in AI safety.  In addition, we are beginning to see conferences 
devoted to contemporary ethical issues raised by machine learning.  See, for example, the workshop 
\href{https://www.aclweb.org/portal/content/first-workshop-ethics-natural-language-processing}{Ethics in 
Natural Language Processing}.}

\subsection{Computer Algebra and Domain-Specific Oracles for Mathematical Computation}
The question answering systems described above all rely on natural language processing to varying 
degrees.  In addition, their domain of applicability has tended towards ``ordinary'' day-to-day knowledge 
useful to a wide array of consumers.  Another type of question answering system is a computer algebra 
system (CAS).  Computer algebra has traditionally referred to systems for computing specific results to 
specific mathematical equations, for example, computing derivatives and integrals, group theoretic 
quantities, etc.  In a sense, we can think of computer algebra as a set of algorithms for performing what 
an applied mathematician or theoretical physicist might work out on paper and pencil.  Indeed, some of 
the early work in computer algebra came from quantum field theory---one of the first computer algebra 
systems was Veltman's \emph{Schoonschip} for performing field theoretic computations that led to the 
theory of electroweak unification \cite{Schoonschip}.  \\

As computer algebra systems have grown in popularity, their functionality has expanded substantially
to cover a wide range of standard computations in mathematics and theoretical physics, including differentiation,
integration, matrix operations, manipulation of symbolic expressions, symbolic substitution, algebraic equation solving,
limit computation, and many others.  Computer algebra systems typically run in a \texttt{read, evaluate, print} loop (\texttt{repl}), 
 and in the research and education context, their popularity has also grown as a result of the notebook model pioneered
by the \emph{Mathematica} system, allowing for computations in CASs to closely mimic the sequential, paper and pencil
work of mathematicians and theoretical physicists.  \\

In assessing the long-term utility of CASs, it is important to note that there is little reason to believe that
computer algebra will be subsumed by other branches of AI research such as machine learning.  Indeed, 
recent research has 
demonstrated applications of machine learning to both computer algebra and theorem proving (which 
we discuss in more detail below), via algorithm selection in the former case \cite{huang2016machine} 
and proof assistance in the latter \cite{irving2016deepmath, komendantskaya2012machine}.  While 
certainly not as visible as machine learning, computer algebra and theorem proving are very much 
active and deep areas of research which are also likely to profit from advances in other fields of 
artificial intelligence, as opposed to being replaced by them \cite{bundy_et_al:DR:2012:3731}.  
On the time horizons on which we are likely to 
see human-level artificial intelligence and beyond, we can expect that these systems will become quite 
powerful, and possess capabilities that 
may be useful in the construction of more general AI systems.  Therefore, it is worth examining such 
systems from the perspective of AI 
safety.

\subsection{Briefly Clarifying Nomenclature}
Before proceeding, we want to explicitly describe issues relating to nomenclature that have arisen in the 
discussion thus far, and state our choices for terminology.  Given that the phrase ``Oracle AI'' has 
become common usage in the AI safety community, we will continue to use this phrase, with the first 
word capitalized, as well as the acronym OAI.  Where clarification is needed, we may also use the full 
phrase ``superintelligent oracle AI,'' without capitalization.  \\

For more modest use cases of the word oracle, we will either refer to ``domain-specific oracles,'' or state the 
domain of knowledge where the oracle is applicable.  We can, at the very least in the abstract, consider 
extending this terminology to other domains such as ``physics oracles,'' ``cell biology oracles,'' or 
``ethics oracles'' and so on.  Therefore, the remainder of the article 
will be concerned with safety and robustness issues in the design of ``math oracles.''

\section{Robust Computer Algebra and Integrated Theorem Proving}\label{itp-cas}
\begin{quote}
{\small \emph{Today we should consider as a standard feature much closer interaction between proof 
assistance and computer algebra software. Several areas can benefit from this, including specification 
of interfaces among components, certification of results and domains of applicability, justification of 
optimizations and, in the other direction, use of efficient algebra in proofs.}\\ -
\textbf{Stephen Watt in \emph{On the future of computer algebra systems at the threshold of 2010}}}
\end{quote}

As we described above, computer algebra systems can be thought of as question answering systems 
for a subset of mathematics.  A related set of systems are interactive proof assistants or interactive 
theorem provers (ITPs).  While ITPs are also systems for computer-assisted mathematics, it is for a 
different 
mathematical context, for computations in which one wishes to construct a proof of a general kind of 
statement.  In other words, rather than computing specific answers to specific questions, ITPs are used 
to show that candidate mathematical structures (or software systems) possess certain properties. \\  

In a sense, the 
distinction between theorem proving and computer algebra should be viewed as a historical anomaly.  
From the perspective of philosophical and logical efforts in the early 20th century that led to the 
``mechanization of mathematics'' the distinction between computing the $n^{th}$ Laguerre polynomial 
and constructing a proof by induction might have been viewed as rather artificial, although with the 
benefit of hindsight we can see that the two types of tasks are quite different in practice 
\cite{beeson2004mechanization}.  \\

The role of ITPs in the research world is very different from that of CASs.  Whereas CASs allow researchers
to perform difficult computations that would be impossible with paper and pencil, constructing proofs using 
ITPs is often more difficult than even the most rigorous methods of pure mathematics.  In broad terms, the 
overhead of using ITPs to formalize theorems arises from the fact that proofs in these systems must proceed
strictly from a set of formalized axioms so that the system can verify each computation.  Consequently, ITPs
(and related systems, such as automatic theorem provers) are largely used for verifying properties of 
mission-critical software systems which require a high-degree of assurance, or for hardware verification,
where mistakes can lead to costly recalls \cite{seL4, kaivola2009replacing, fix2008fifteen, kern1999formal, Kropf}.  \\

\begin{figure*}[h]
\begin{center}
\includegraphics[width=\textwidth]{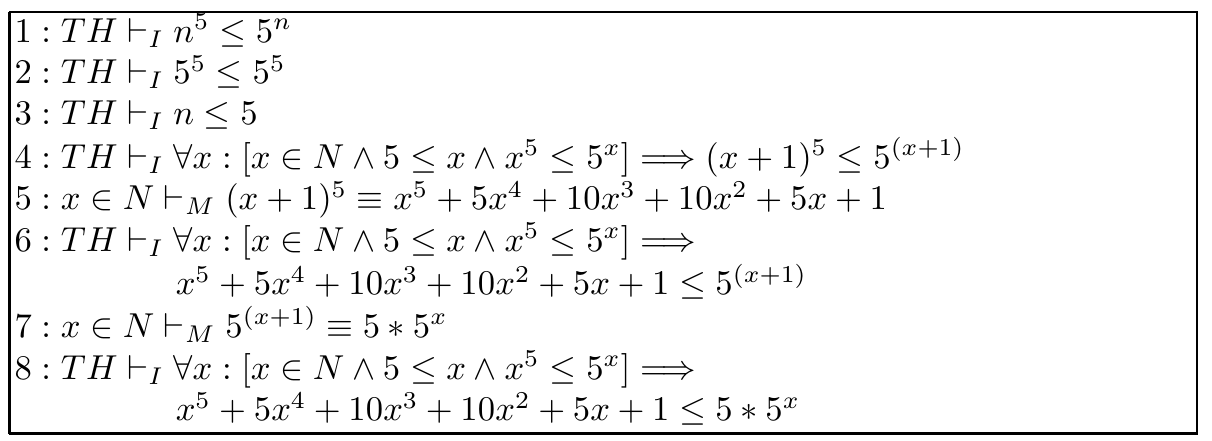}
\caption{\label{fig:maple_isabelle}Example of a polynomial identity proven by integrating the Maple 
computer algebra system with Isabelle.  Maple's simplifier is used for expanding polynomials---a 
powerful complement to the theorem proving architecture of Isabelle which allows for the setup of a 
proof by induction.}
\end{center}
\end{figure*}

As the quotation above suggests, many academic researchers view the integration of interactive proof 
assistants and computer algebra systems as desirable, and there have been numerous efforts over the 
years at exploring possible avenues for achieving this objective \cite{Ballarin, HOLCAS, Watt, 
Theorema} (a more complete list is given below).  By integrating theorem proving with computer 
algebra, we would be opening up a wealth of potentially interoperable algorithms that have to date 
remained largely unintegrated.  To cite one such example, in \cite{MapleIsabelle}, the authors have 
developed a framework for exchange of information between the Maple computer algebra system and 
the Isabelle interactive theorem prover. They show a simple problem involving the proof of an 
elementary polynomial identity that could be solved with the combined system, but in neither system 
alone (see Fig. \ref{fig:maple_isabelle}). \\

We cite this example to demonstrate how a simply stated elementary problem cannot be solved in 
existing environments for either computer algebra or proof assistance.  The computer algebra system 
does not have the capacity for structural induction and theorem provers generally have rather weak 
expression simplifiers.  There are numerous examples such as this one in the academic literature. \\

Another key difference between CASs and ITPs is the architectural soundness of the respective 
systems.  As we will discuss below, computer algebra systems have well-defined architectural 
deficiencies, which while not a practical issue for the vast majority of use cases, pose problems for their 
integration with theorem provers, which by their nature, are designed to be architecturally sound.  In the 
context of superintelligent AI systems, the architectural problems of CASs are potential points of 
weakness that could be exploited for malicious purposes or simply lead to unintended and detrimental consequences.  
Therefore, we use the phrase ``robust 
computer algebra'' to refer to CASs which lack the problems 
that have been identified in the research literature.  In the section below, we combine the discussion 
of robust computer algebra and integration with interactive theorem provers, as there is a spectrum of 
approaches which address both of these issues to varying degrees.  

\subsection{A Taxonomy of Approaches}
There are many possible avenues to tackle the integration of theorem provers with computer algebra 
systems.  We give 4 broad categories characterizing such integration efforts\footnote{This classification 
was first described by Kaliszyk and Wiedijk \cite{HOLCAS} in a paper arguing for an architecture which 
we list as the fourth category given above.}: 

\begin{enumerate}
\item \textbf{Theorem provers built on top of computer algebra systems:} These include Analytica, 
Theorema, RedLog, and logical extensions to the Axiom system \cite{clarke1992analytica, Theorema, 
dolzmann1997redlog, jenks2013axiomtm, poll1998adding}  .
\item \textbf{Frameworks for mathematical exchange between the two systems:} This category includes 
MathML, OpenMath, OMSCS, MathScheme, and Logic Broker \cite{miner2005importance, 
buswell2004open, calmet2004toward, carette2011mathscheme, armando2000towards}. 
\item \textbf{``Bridges'' or ``ad-hoc'' information exchange solutions:} The pairs of systems in this 
category include bridges combining PVS, HOL, or Isabelle with Maple, NuPRL with Weyl, Omega with 
Maple/GAP, Isabelle with Summit, and most recently, Lean with \emph{Mathematica} 
\cite{MapleIsabelle, adams2001computer, harrison1998skeptic, 
ballarin1995theorems, jackson1994exploring, siekmann2002proof, ballarin1999pragmatic, 
LeanMathematica2017}.  The 
example given above, bridging Isabelle and Maple, is an example of an approach from this category.
\item \textbf{Embedding a computer algebra system inside a proof assistant:} This is the approach 
taken by Kaliszyk and Wiedijk in the HOLCAS system.  In their system, all expressions have precise 
semantics, and the proof assistant proves the correctness of each simplification made by the computer 
algebra system \cite{HOLCAS}.
\end{enumerate}

One primary aspect of integration that differentiates these approaches is the degree of trust the 
theorem prover places in the computer algebra system.  Computer algebra systems give the false 
impression of being monolithic systems with globally well-defined semantics.  In reality, they are large 
collections of algorithms which are neatly packaged into a unified interface.  Consequently, there are 
often corner cases where the lack of precise semantics can lead to erroneous solutions.  Consider the 
following example:  

\begin{figure}[h]
\begin{center}
\frame{\includegraphics[scale=.32]{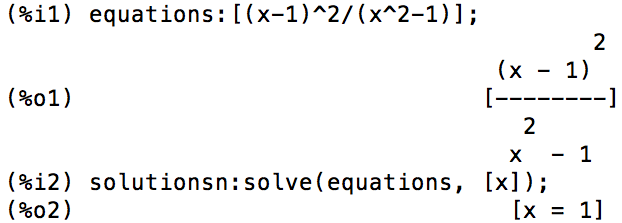}}
\caption{\label{fig:solve-error-simp}Example of an incorrect solution to a simple polynomial equation by 
a computer algebra system.}
\end{center}
\end{figure}

The system incorrectly gives 1 as a solution, even though the given polynomial has an indeterminate 
value for $x = 1$.  However, because the expression is treated as a fraction of polynomials, it is first 
simplified before the solve operation is applied.  In other words, there is an unclear semantics between 
the solver module and the simplifier which leads to an incorrect result.  \\

Another simple example is the following integral:
\begin{figure}[h]
\begin{center}
\includegraphics[scale=.70]{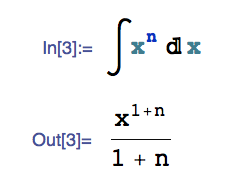}
\caption{\label{fig:solve-error-noncomm}A problem arising in symbolic integration due to the non-commutativity 
of evaluation and substitution.}
\end{center}
\end{figure}

Making the substitution $n = -1$ gives an indeterminate result, while it is clear by inspection that the 
solution to the integral for $n = -1$ is simply $ln(x)$.  This belongs to a class of problems known as the 
\emph{specialization problem}, namely that expression evaluation and variable substitution do not 
commute \cite{Ballarin}.  So while we have seen above that theorem proving can benefit tremendously 
from the wealth of algorithms for expression simplification and mathematical knowledge in computer 
algebra, there is the potential cost of compromising the reliability of the combined system.  As a possible
application to current research in AI safety, consider the decision-theoretic research agenda 
for the development of safe, superintelligent AI systems outlined in \cite{yudkowsky2013tiling, lavictoire2015introduction, 
barasz2014robust, fallenstein2014problems, soares2015toward}.  If we require formal guarantees of 
correctness at any point in a sequence of computations in which computer algebra is used, current systems 
would be unable to provide the necessary framework for constructing such a proof.

\subsubsection{Qualitatively Certified Computations}  
In our taxonomy of approaches to bridging theorem provers with computer algebra, we described how a 
key distinction was the degree of trust that the theorem prover places in the computer algebra system.  
For instance, approaches which build theorem provers on top of computer algebra systems do not 
address the architectural issues with CASs.  They are integrative, but not more sound.  On the other 
extreme, building a computer algebra system on top of a theorem prover allows for a degree of trust 
that is on par with that of the theorem prover itself.  However, this approach has the distinct 
disadvantage that computer algebra systems represent many hundred man-years worth of effort. \\

The more intermediate approaches involving common languages for symbolic exchange or ad-hoc bridges, 
bring to light an important notion in the spectrum of provable safety, namely 
the ability to assign probabilities for the correctness of computations.  In \cite{garrabrant2016logical}, 
the authors present an algorithm for assigning probabilities to any statement in a formal language.  We 
might ask what strategies might look like that have a similar goal in mind, but are significantly weaker.  
Interfaces between theorem provers and computer algebra systems provide a concrete example where 
we can ask a question along these lines.  Fundamentally, in such an interface, the computer algebra 
system is the weaker link and should decrease our confidence in the final result.  But by how much?  
For instance, in the example given in Figure \ref{fig:maple_isabelle}, how should we revise our 
confidence in the result knowing that polynomial simplification was conducted within a computer algebra 
system? \\

It is worth asking for simple answers to this question that do not require major theoretical advances to 
be made.  For instance, we might imagine curating information from computer algebra experts about 
known weaknesses, and use this information to simply give a qualitative degree of confidence in a 
given result.  Or, for example, in a repository of formal proofs generated using integrated systems, steps 
of the proof that require computer algebra can be flagged and also assigned a qualitative measure of 
uncertainty.  \\

The relationship that this highly informal method of giving qualitative certification to computations has 
with the formal algorithm developed in \cite{garrabrant2016logical} can be compared to existing 
techniques in the software industry for ensuring correctness.  On the one hand, unit testing is a 
theoretically trivial, yet quite powerful practice, something along the lines of automated checklists for 
software.  The complexities of modern software would be impossible to handle without extensive 
software testing frameworks \cite{Beck2002, Osherove2013, maximilien2003assessing, 
erdogmus2005effectiveness, sarma2016unit}.  On the other hand, formal verification can provide substantially stronger 
guarantees, yet is a major undertaking, and the correctness proofs are often significantly more 
demanding to construct than the software itself.  Consequently, as discussed in Section \ref{itp-cas},
formal verification is much less frequently used in industry, typically only in exceptional circumstances 
where high guarantees of correctness are required, or for 
hardware verification \cite{seL4, kaivola2009replacing, fix2008fifteen, kern1999formal, Kropf}.  \\

Integrated systems for computer algebra and theorem proving give rise to a quite interesting (and 
perhaps ironic) opportunity to pursue simple strategies for giving qualitative estimates for the 
correctness of a computation.

\subsubsection{Logical Failures and Error Propagation}
As the examples described above demonstrate, errors in initial 
calculations may very well propagate and give rise to non-sensical results.  As AI systems capable of performing
mathematical computation become increasingly sophisticated and embedded as part of design workflows
for science and engineering (beyond what we see today), we could imagine such errors being quite costly
and difficult to debug.  In the case of a 
superintelligent AI system, more concerning scenarios would be if systematic errors in computer 
algebra could be exploited for adversarial purposes or if they led to unintentional accidents on a large scale.\\

The issue of error propagation is another example of a concrete context for pursuing simple strategies 
for assigning qualitative measures of certainty to computations performed by integrated theorem 
proving / computer algebra systems.  For instance, we may be less inclined to trust a result in which the 
computer algebra system was invoked early on in a computation as opposed to later.  With curated data 
from computer algebra experts on the reliability or failure modes of various algorithms, we might also 
chain together these informal estimates to arrive at a single global qualitative estimate.  If multiple 
systems were to be developed independently, or which were based on fundamentally different 
architectures, we might also be significantly more confident in a result which could be verified by two 
separate systems.  

\subsubsection{Additional Topics}
Some related ideas merit investigation in the broader context of mathematical computation:
\begin{itemize}
\item \textbf{Integrating SMT solvers with interactive theorem provers:} Satisfiability modulo theories 
(SMT) solvers are an important element of automated reasoning and there have been efforts 
analogous to those described above to bridge SMT solvers with interactive theorem provers 
\cite{keller2013matter, armand2011modular}.  \\
\item \textbf{Identifying the most important / widely used algorithms in computer algebra:}  Computer 
algebra systems have grown to become massive collections of algorithms extending into domains well 
outside of the realm of mathematics.  If the purely mathematical capacities of CASs prove to be 
useful in future AI systems, it would be valuable to rank order algorithms by their popularity or 
importance.\\

One approach would be to do basic textual analysis of the source code from GitHub or StackExchange.  
This would also allow for more targeted efforts to directly address the issues with soundness in core 
algorithms such as expression simplification or integration.  In the context of the HOLCAS system 
described above, for example, it would be valuable to have rough estimates for the number of man-hours 
required to implement a minimal CAS with the most widely used functionality on top of a theorem 
prover.  \\
\item \textbf{Proof checkers for integrated systems:}
Proof checkers are important tools in the landscape of formal verification and theorem proving.  Indeed, 
as it is often much less computationally expensive to verify the correctness of a proof than to generate it 
from scratch, the availability of proof checkers for the widely used interactive theorem provers is one 
reason we can be confident in the correctness of formal proofs 
\cite{harrison2006towards,pollack1998believe}. \\

As we described above, strategies for integrating computer algebra with theorem provers can 
potentially result in a combined system which is less trustworthy than the theorem prover alone.  
Therefore, the availability of proof checkers for combined systems would be a valuable resource in 
verifying proof correctness, and in certain mathematical domains, potentially provide an avenue for 
surmounting the need to directly make the CAS itself more architecturally robust. \\

The development of integrated proof checkers is likely to be a substantial undertaking and require novel 
architectures for integrating the core CAS and ITP systems distinct from what has been described 
above.  However, it is a largely unexplored topic that merits further investigation.  \\

\item \textbf{Analyzing scaling properties of algorithms for computer algebra and theorem proving as a 
function of hardware resources:}
The premise of the analysis presented above is that CASs (and integrated theorem proving) are likely to 
remain sufficiently architecturally stable and useful on a several decade time-horizon in the construction 
of AI 
systems.  On the other hand, as we argued earlier, it is much less clear that the same will be true of the 
most visible, NLP-based, consumer-oriented question answering systems.  To make these arguments 
more rigorous, it would be valuable to develop quantitative predictions of what the capabilities will be of 
existing algorithms for computer algebra and theorem proving when provided with substantially 
expanded hardware resources.  For instance, we might examine problems in mathematics or theoretical physics for 
which na\"{i}ve solutions in CASs are intractable with current resources, but which may be feasible with 
future hardware.  \\

\item \textbf{The cognitive science of computer algebra:}
What role has computer algebra played in theoretical physics and mathematics?  How has it influenced 
the thinking process of researchers?  Has computer algebra simply been a convenience that has shifted 
the way problems are solved, or has it fundamentally enabled new problems to be solved that would 
have been completely intractable otherwise?  \\

The cognitive science of mathematical thought is a substantial topic which overlaps with many 
established areas of research \cite{hardy1946psychology, dehaene2011number, drijvers2005computer, 
drijvers2002learning, lakoff2000mathematics}.  However, a systematic review of research in mathematics and theoretical 
physics since the advent of computer algebra and its role in the mathematical thought process is an 
underexplored topic.  It would be an interesting avenue to pursue in understanding the role that CASs, 
ITPs, and integrated systems may come to play in superintelligence, particularly in the case of neuromorphic
systems that have been modeled after human cognition.  These questions also relate to 
understanding the scaling properties of CAS and theorem proving algorithms 
as well as cataloguing the most widely used algorithms in computer algebra.  

\end{itemize}

\section{Conclusion}
The aim of this article has been to examine pre-existing research objectives in computer science and 
related disciplines which align with problems relevant to AI safety, thereby providing concrete, practical 
context for problems which are otherwise of a longer time horizon than most research.  In particular, we 
focused on the notion of ``Oracle AI'' as used in the AI safety community, and observed that the word 
oracle has two meanings in the context of superintelligent AI systems.  One usage refers to a 
subsystem of a larger AI system queried for domain-specific tasks, and the other to superintelligent AI 
systems restricted to only answer questions.  \\

We examined contemporary question answering systems (QASs) and argued that due to their 
architectural heterogeneity, consumer-oriented, NLP-based systems do not readily lend themselves to 
rigorous analysis from an AI safety perspective.  On the other hand, we identified computer algebra 
systems (CASs) as concrete, if primitive, examples of domain-specific oracles.  We examined well-known architectural
deficiencies with CASs identified by the theorem proving community and argued that the integration of 
interactive theorem provers (ITPs) with CASs, an objective that has been an area of research in the 
respective communities for several decades, provides a set of research problems and practical software 
projects related to the development of powerful and robust math oracles on a multi-decade time horizon.  
Independent of their role as domain-specific oracles, such systems may also prove to be useful tools for 
AI safety researchers in proving the functional correctness of other components of an AI architecture.  
Natural choices of systems to use would be interfaces for the Wolfram Language, the most widely 
used computer algebra system, with one of the HOL family of theorem provers or Coq, 
both of which have substantial repositories of formalized proofs 
\cite{wolfram2015elementary, paulson1989foundation, paulson1994isabelle, bertot2013interactive}, 
or a more modern ITP such as Lean \cite{de2015lean, LeanMathematica2017}.     \\

Rather than representing a bold and profound new agenda, we view these projects as being
concrete and achievable goals that may pave the way to more substantial research directions.  
Because the topics we have discussed have a long and rich academic history, there are a number 
of ``shovel-ready'' projects appropriate for students anywhere from undergraduates to PhD students 
and beyond.  Good undergraduate research projects would probably start with some basic data science 
to catalogue core computer algebra algorithms by their usage and popularity.  From there, it would be 
useful to have an estimate of what certified implementations of these algorithms would entail, 
whether formally verified implementations, or along the lines of Kaliszyk and Wiedijk's HOLCAS 
system where the CAS is built on top of a theorem prover.  Also useful would 
be a systematic study of role that computer algebra has played in mathematics and theoretical physics.  
This would have some interesting overlap with cognitive psychology, and these three projects 
together would make for an approachable undergraduate thesis, or a beginning project for a 
graduate student.  A solid PhD thesis devoted to the topic of Oracle AI might involve tackling 
approaches to oracles stemming from reinforcement learning (RL) \cite{armstrong2016safely, armstrong2017},
as well as more advanced theorem proving and CAS related topics such as investigating 
the development of a hybrid architecture that would allow for proof-checking.  
A student who worked on these projects for several years would develop a unique 
skill set spanning philosophy, machine learning, theorem proving, and computer algebra.  \\

In the context of superintelligent oracle AIs which may possess the ability to manipulate
a human user, we differentiate between addressing architectural 
or algorithmic deficiencies in subsystems versus general control methods or containment strategies.  
Given that strong mathematical capabilities
are likely to be useful in the construction of more general AI systems, designing robust CASs 
(and any other domain-specific oracle)
is an important counterpart to general control strategies, as the top-level AI system will have fewer loopholes to exploit.  
Controlling OAIs poses a distinct set of challenges for which concrete mathematical 
analysis is in its infancy \cite{armstrong2016safely, armstrong2017, armstrong2017oracle}.  Nonetheless, considering 
how little attention has been given to the superintelligence control problem in general, we are optimistic 
about the potential to translate the high-level analyses of OAIs that have arisen in the AI safety 
community into the mathematical and software frameworks of modern artificial intelligence.   

\section*{Acknowledgements}
We would like to thank Stuart Armstrong, David Kristoffersson, Marcello Herreshoff, 
Miles Brundage, Eric Drexler, Cristian Calude, and several anonymous reviewers 
for insightful discussions and feedback on the manuscript.  We would also like to thank
the guest editors of \emph{Informatica}, Ryan Carey, Matthijs Maas, Nell Watson,
and Roman Yampolskiy, for organizing this special issue.  

\bibliographystyle{ieeetr}
\bibliography{computer_algebra_theorem_proving_AI}

\end{document}